\pdfoutput=1

\documentclass[11pt]{article}

\usepackage[final]{coling}

\usepackage{times}
\usepackage{latexsym}
\usepackage{multirow}
\usepackage{multicol}
\usepackage{svg}
\usepackage{booktabs} 
\usepackage{tcolorbox}
\usepackage{graphicx}
\usepackage{lipsum}
\usepackage[inline]{enumitem}
\usepackage{amsmath}
\usepackage{adjustbox}

\usepackage[T1]{fontenc}

\usepackage[utf8]{inputenc}

\usepackage{microtype}

\usepackage{inconsolata}

\usepackage{graphicx}

\title{\textit{``Stupid robot, I want to speak to a human!''}

User Frustration Detection in Task-Oriented Dialog Systems}

\author{
    Mireia Hernandez Caralt$^*$, 
    Ivan Sekuli\'c$^*$,
    Filip Carevi\'c$^*$, 
    Nghia Khau, \\
    \bf  Diana Nicoleta Popa, 
    Bruna Guedes, 
    Victor Guimarães,
    Zeyu Yang, \\
    \bf Andre Manso,
    Meghana Reddy,
    Paolo Rosso,
    Roland Mathis
\\
        Telepathy Labs GmbH, Zürich, Switzerland \\ \texttt{
        \{firstname\}.\{lastname\}@telepathy.ai}}

\begin{document}
\maketitle
\begin{abstract}
Detecting user frustration in modern-day task-oriented dialog (TOD) systems is imperative for maintaining overall user satisfaction, engagement, and retention.
However, most recent research is focused on sentiment and emotion detection in academic settings, thus failing to fully encapsulate implications of real-world user data.
To mitigate this gap, in this work, we focus on user frustration in a deployed TOD system, assessing the feasibility of out-of-the-box solutions for user frustration detection.
Specifically, we compare the performance of our deployed keyword-based approach, open-source approaches to sentiment analysis, dialog breakdown detection methods, and emerging in-context learning LLM-based detection.
Our analysis highlights the limitations of open-source methods for real-world frustration detection, while demonstrating the superior performance of the LLM-based approach, achieving a 16\% relative improvement in F1 score on an internal benchmark.
Finally, we analyze advantages and limitations of our methods and provide an insight into user frustration detection task for industry practitioners.
\end{abstract}

\def\thefootnote{*}\footnotetext{These authors contributed equally to this work.}\def\thefootnote{\arabic{footnote}}

\section{Introduction}

\citet{berkowitz1989frustration} defines frustration as an emotional state that is a result of the occurrence of an obstacle that prevents the satisfaction of a need.
As such, in the context of task-oriented dialog (TOD) systems, detection of user's frustration is an essential component in ensuring the fulfillment of the user's goal~\cite{hinrichs2018}.
The importance stems from the fact that frustrated users often abruptly terminate their conversation with a TOD system, leading to a low likelihood of their return.
Thus, timely detection of user frustration has many benefits, as the system can employ dialog flow repair techniques or transfer the user to a human agent, in order to improve the user experience~\cite{zhang-etal-2023-groundialog}.

While a large body of work on the topic of emotion detection in dialog~\cite{pereira2022, Zhang2023DialogueLLMCA, wang-etal-2024-emotion} and dialog breakdown detection~\cite{li2020,terragni-etal-2022-betold} exists, research on user frustration detection in real-world TOD systems is scarce.
Therefore, in this paper, we present a unique perspective from the industry, analyzing user frustration in conversations from a deployed TOD system.

\begin{figure}[t]
 \centering
   \includegraphics[width=\linewidth]{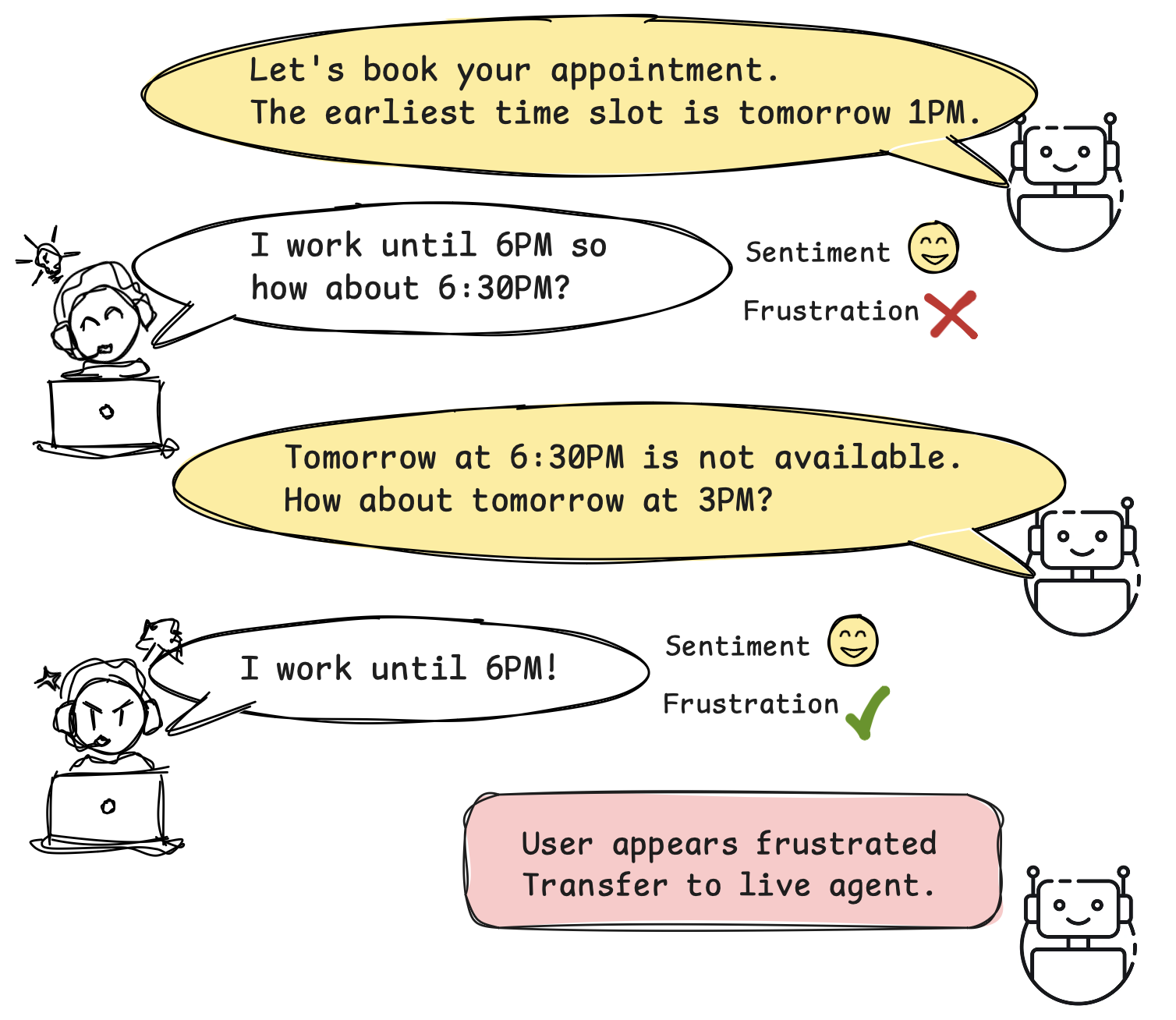}
\caption{Example of user frustration in a deployed TOD system. The user can only come after 6PM due to work, but the system misses this and suggests the next available slot. Traditional sentiment models often fail to detect such nuances, as there is no explicit mention of negative sentiment.}
\label{fig:motivating_figure}
\end{figure}

Specifically, we showcase the gap between research-oriented approaches and real-world applications. 
To this end, we compare emotion detection datasets constructed for academic research, namely EmoWoZ~\cite{feng-etal-2022-emowoz}, to our internal data gathered from real users conversing with a deployed TOD system, finding several differences, discussed in Sect.~\ref{sec:analysis:academic}.
We hypothesize that the differences arise mainly from the fact that there is a real sense of urgency and importance in completing the real-world tasks, whereas users in academic benchmarks are more cooperative and may even tolerate mistakes from the TOD system.

Additionally, we assess established methods for sentiment analysis~\cite{hartmann2023more}, emotion detection~\cite{huang2022distilbert}, and dialog breakdown detection~\cite{bodigutla-etal-2020-joint} on our internal dataset, concluding they are mostly insufficient for successfully detecting user frustration.
In an attempt to mitigate this gap, we propose two approaches, stemming from two different angles: 
\begin{enumerate*}[label=\roman*)]
    \item currently deployed keyword-based user frustration detection method, grounded in sentiment analysis;
    \item novel, emerging in-context learning LLM-based method.
\end{enumerate*}

We conclude that our rule-based approach, although precise, fails to detect user frustration in a large number of cases.
Furthermore, LLM-based approach outperforms all of the aforementioned approaches on our internal frustration detection benchmark. 
Finally, we outline promising directions for future work through the industry-relevant perspective.


\section{Related Work}
\label{sec:rw}
Early detection of user frustration is essential for improving the quality of TOD systems. 
 Causes of user frustration  include poor performance, poor utility and poor usability of the systems they interact with~\cite{Hertzum2023}, such as the inability of the system to correctly understand user requests, a mismatch between user expectations and obtained results or an overall dissatisfaction with the provided results.

Much of the existing work on user frustration explores the problem from a non-technical view~\cite{Goetsu2020, brendel2020you, Hertzum2023}, focuses on the 
broader scope of human-machine interaction 
~\cite{weidemann2021} or explores mitigating breakdowns in such interactions~\cite{li2020, terragni-etal-2022-betold}, yet without explicitly targeting user frustration detection. Additionally, some studies have focused on  user satisfaction estimation and overall dialog quality assessment~\cite{rach-etal-2017-interaction, bodigutla2019, bodigutla-etal-2020-joint, sun2021}
Since determining user frustration is a more targeted goal, it can be seen as a subset of such studies, making the nuances of mapping user satisfaction levels to frustration both challenging and error-prone. 
Similarly, emotion detection~\cite{pereira2022, Zhang2023DialogueLLMCA,wang-etal-2024-emotion} could be seen as an important prerequisite for user frustration assessment. The main idea behind such studies is that frustration is present if emotions such as dissatisfaction or anger can be detected in the given textual content. 

As detecting user frustration is both subtle and complex, hand-crafted feature engineering may also not be enough~\cite{hinrichs2018, ang2002}. Meanwhile, LLMs have recently yielded impressive performance on a variety of tasks, encoding much general world knowledge~\cite{Touvron2023LLaMAOA, OpenAI_GPT4_2023}. We thus investigate to what extent such an approach could be suitable for detecting user frustration. To the best of our knowledge, this is the first work targeting unsupervised user frustration detection using LLMs within deployed TOD systems.

\section{User Frustration Detection}
\label{sec:method}
In this section, we first formalize the task of user frustration detection.
Next, we describe our deployed rule-based approach and propose a novel method based in in-context learning with LLMs.
Finally, we describe competitive baselines used as comparison to our approaches and details regarding our internal benchmark data, sourced from a deployed TOD system.


We define $\mathcal{U}$ as the domain of all textual utterances. Then, given an ordered list of tuples (i.e., a dialog history) $H = [(s_i, u_i)\mid i\in \{1, \dots, t\}]$, where $s_i \in \mathcal{U}$ and $u_i \in \mathcal{U}$ denote system and user utterance at dialog turn $i$, respectively, the goal is to find such function $f: \mathcal{U} \times \mathcal{U} \rightarrow \{0,1\} $ that for presence of frustration in the dialog outputs positive label, and negative otherwise.


\subsection{Rule-Based Approach}
\label{sec:method:keyword}
Our deployed user frustration detection system relies on keyword match in user utterances.
Specifically, we have curated a set of keywords $K = \{kw_1,\dots,kw_N\}$, mainly composed of profanity, words explicitly indicating negative emotion, and insults, indicating potential frustration.
In practice, for each keyword $kw_i \in K$, we check if it appears in the current user utterance $kw_i \in u_t$; if a match is found, the conversation is marked as frustrated.



\subsection{LLM-Based Approach}
\label{sec:method:llm}
In-context learning (ICL) paradigm with LLMs has demonstrated strong performance across a wide range of tasks, without the need for time- and compute-expensive fine-tuning~\cite{brown2020languagemodelsfewshotlearners}.
We hypothesize that LLMs are well-suited to identifying the nuanced indicators of user frustration when given an appropriate description and context. Therefore, we propose a novel approach for frustration detection in TOD systems, based on ICL with LLMs.

Specifically, we design an ICL prompt $\mathcal{P}(T,D,H)$ which includes: \textbf{i) task description (T)} of user frustration in the context of TOD systems and common cues for its identification, \textbf{ii) domain (D)} of the conversation (e.g., booking appointments), which helps the language model understand the expected interaction patterns and \textbf{iii) conversation history (H)}  formatted as a string with \emph{``USER:''} and \emph{``SYSTEM:''} prefixes to distinguish between roles. 
The LLM processes this context and generates the corresponding binary frustration label
$UF = f_{LLM}(\mathcal{P}(T, D, H))$.

The prompt strings used are detailed in App.~\ref{app:prompt}. In the experiments shown in Sect. \ref{sec:results} we also provide the results obtained when further augmenting the prompt with few-shot examples, and compare them with this zero-shot approach.




\begin{table*}[h!]
\centering
\small
\begin{tabular}{lllllllr}
\toprule
                                                              & \multicolumn{3}{c}{\textbf{UF = 0 (not frustrated)}} & \multicolumn{3}{c}{\textbf{UF = 1 (frustrated)}}  &                  \\
                                                              & P   & R   & $F_1$   & P   & R   & $F_1$  & Macro-$F_1$\\

\midrule
\textbf{Sentiment Analysis}~\cite{hartmann2023more}                       &                  &                 &               &                &                  &                 &                  \\
\texttt{RoBERTa-Sent-FC}                           & 0.73           & 0.11           & 0.18        & 0.34         & 0.92           & 0.50          & 0.34            \\
\texttt{RoBERTa-Sent-LU}                      & 0.77           & 0.72           & 0.75        & 0.51         & 0.58           & 0.54          & 0.65           \\

\textbf{Emotion Detection}                      &                  &                 &               &                &                  &                 &                  \\
\texttt{DistilBERT-EmoWoZ-FC}~\cite{huang2022distilbert}                              & 0.70           & 0.77           & 0.73        & 0.42         & 0.34           & 0.38          & 0.56           \\
\texttt{DistilBERT-EmoWoZ-LU}                         & 0.74           & 0.68           & 0.71        & 0.45         & 0.52           & 0.48          & 0.60           \\

\texttt{DistilRoBERTa-Emo-FC}~\cite{hartmann2022emotionenglish}                      & 0.67           & 1.00           & 0.80        & 0.00         & 0.00           & 0.00          & 0.40           \\
 \texttt{DistilRoBERTa-Emo-LU}                 & 0.68           & 1.00           & 0.81        & 0.88         & 0.04           & 0.07          & 0.44           \\
\textbf{Dialog Breakdown Detection}                &                  &                 &               &                &                  &                 &                  \\
\texttt{DBD+LogReg}~\cite{bodigutla-etal-2020-joint}                  & 0.78           & 0.93           & 0.85        & 0.78         & 0.46           & 0.58          & 0.71           \\
\textbf{Rule-Based Approach}                                 &                  &                 &               &                &                  &                 &                  \\
\texttt{Keyword Matching}                           & 0.67           & \textbf{1.00}           & 0.80        & \textbf{1.00}          & 0.01           & 0.01           & 0.41           \\
\textbf{LLM-based ICL Approach}                    &                  &                 &               &                &                  &                 &                  \\
\texttt{GPT-4o-zero-shot}                                               & 0.98  & 0.83  & 0.90  & 0.74  & 0.96  & 0.83  & 0.86\\
\texttt{GPT-4o-two-shot} & 0.85 &	0.97&	\textbf{0.91}	&0.92	&0.66	&0.77	&0.84	\\
\texttt{Llama-3.1-405B-zero-shot}                                               & \textbf{0.99}  & 0.74  & 0.85  & 0.67  & \textbf{0.99}  & 0.79  & 0.83\\
\texttt{Llama-3.1-405B-two-shot}                                               & 0.97	 & 0.84 & 	0.90 & 	0.75	 & 0.96 & 	\textbf{0.84} & 	 \textbf{0.87}\\
\bottomrule
\end{tabular}
\caption{Results of various approaches for user frustration detection on our deployed TOD system benchmark.}
\label{tab:results}
\end{table*}

\subsection{Baselines}
\label{sec:method:baselines}
We compare our aforementioned in-house methods for user frustration detection with baselines from three different streams of approaches:
\begin{enumerate*}[label=\roman*)]
    \item sentiment analysis;
    \item emotion detection;
    \item dialog breakdown detection (DBD).
\end{enumerate*}

\paragraph{Sentiment Analysis and Emotion Detection.} Sentiment analysis baselines aim to classify the sentiment of a text as either \emph{positive} or \emph{negative}. All samples having \emph{negative} sentiment are considered as entailing user frustration. As the representative of this group, we leverage \emph{RoBERTa-large} model, fine-tuned on a large variety of sentiment classification datasets~\cite{hartmann2023more}. We dub this method \emph{RoBERTa-Sent}. On the other hand, emotion detection involves identifying a specific emotion from a predefined set in a given text. We employ two models: a distilled version of \emph{BERT} model trained on the conversational EmoWOZ~\cite{feng-etal-2022-emowoz} dataset, and and a distilled version of RoBERTA model trained on various emotion classification datasets~\cite{hartmann2022emotionenglish}. The first model predicts the emotions of \emph{satisfaction}, 
 \emph{dissatisfaction}, \emph{abuse}, \emph{apology}, \emph{excitement}, \emph{fear} and \emph{neutrality}, while the latter one classifies the states of \emph{anger}, \emph{fear}, \emph{disgust}, \emph{joy}, \emph{neutrality}, \emph{sadness} and \emph{surprise}. We dub these models \emph{DistilBERT-EmoWoZ} and \emph{DistilRoBERTa--Emo}, respectively.
 In this work, \emph{dissatisfaction} and \emph{abuse}, as well as  \emph{anger} and \emph{disgust} are considered as indicators of users' frustration.
 
 As the above-described methods might be designed for either single-sentence or full-conversation input formats, following \citet{feng-etal-2022-emowoz}, we evaluate them with two different input types:
\begin{enumerate*}[label=\roman*)]
    \item only the last user utterance $u_N$ as input (\emph{-LU});
    \item full conversation $H$ as input (\emph{-FC}).
\end{enumerate*}
The input format is indicated by appending the abbreviation to the method dub.



\paragraph{Feature-Based Dialog Breakdown Detection.}
This method leverages hand-crafted features in order to estimate the amount of user satisfaction on different conversation levels~\cite{bodigutla2019,DBLP:journals/corr/abs-1908-07064}. We adapt this baseline to our use-case by applying a simple classifier on top of the subset of features presented in~\cite{bodigutla2019}. The list of used features is portrayed in App.~\ref{dialogbreakdownappendix}.


\subsection{Data}
Our data consists of real user conversations with our currently deployed TOD system.
The conversations generated by this system are often lengthy and span multiple dialog phases. 
In this work, we focus on two specific dialog phases that are particularly prone to user frustration:
\begin{enumerate*}[label=\roman*)]
    \item booking negotiations, where the system attempts to schedule a suitable time slot for a user seeking an appointment; and 
    \item receptionist, where the system attempts to route the user to the appropriate department or agent based on their needs.
\end{enumerate*}


We collect conversations with more than one turn from a week of production data, resulting in a dataset of 270 booking negotiations and 285 receptionist transfers. Further details and comparison to EmoWoz \cite{feng-etal-2022-emowoz} is shown in Table \ref{tab:dialog_features}.



Although previous work has explored automated signals to detect user frustration, such as hang-ups or requests for a live operator \cite{terragni-etal-2022-betold}, these methods are susceptible to noise, since a frustrated user may choose to continue the conversation, or a user might hang up for reasons unrelated to frustration. Therefore, we conduct manual annotation of the collected data: we employ three in-house experts to annotate each sample with a binary label indicating whether the user is frustrated or not. 
The annotators were provided with guidelines, reaching an inter-rater agreement, as measured by Fleiss' $\kappa$, of $0.48$. This agreement indicates moderate reliability~\cite{landis1977measurement}, while also suggesting a degree of subjectivity in the annotation task.
Disagreements were manually resolved in a post-processing phase. 


\section{Results}
\label{sec:results}

Table~\ref{tab:results} presents the results of our experiments.
While we primarily focus on macro-F1 for performance comparison, we additionally focus on recall for UF=1 (frustrated), since it highlights the proportion of frustrated conversations correctly identified, which has the greatest impact on user satisfaction.
We make several observations from the results, with a follow-up discussion presented in Sect.~\ref{sec:discussion}.

Sentiment analysis models and emotion detection models fine-tuned for the TOD domain perform poorly compared to dialog breakdown or LLM-based ICL approaches. Notably, these baselines perform the best when only the final user utterance is considered, with the performance of the best model dropping from 66\% to 34\% when analyzing the full conversation history. This suggests that these models capture only the emotion in isolated utterances, but fail to detect frustration cues embedded in the broader conversational context.

Features derived from the dialog breakdown detection domain are effective for detecting user frustration, outperforming  sentiment and emotion detection baselines. However, although DBD outperforms them in terms of Macro-F1, it does seem inclined towards the UF=0 class, as suggested by a relatively low recall in the UF=1 class. 

Our currently deployed keyword-based approach achieves 100\% precision, but suffers from an extremely low recall of only 1\% for frustrated conversations, resulting in a very low Macro-F1 score of 41\%. This indicates that poor conversation handling does not always manifest as overtly negative language. While keyword-based methods may be inexpensive, they are inadequate for capturing the full range of frustrated scenarios.

ICL with LLMs outperforms all other approaches, both in zero- and two-shot settings. Specifically, in terms of Macro-F1, we observe more that $+33\%$ relative improvement over sentiment and emotion detection methods and $+22\%$ relative improvement over the DBD method. We further note the comparable performance of both \texttt{LLaMA-3.1-405B} and \texttt{GPT-4o} in both zero- and few-shot settings, suggesting that our ICL prompt generalizes well across different LLMs and number of shots. However, adding few-shot examples to the prompt does not yield substantial performance improvement, and even slightly degrades performance for GPT-4o. 

\begin{table}[h!]
\centering
\small
\begin{tabular}{lr}
\toprule
                                                              & Macro-$F_1$\\

\midrule
\textbf{Sentiment Analysis}~\cite{hartmann2023more} &                  \\
\texttt{RoBERTa-Sent-FC}          &      0.34            \\
\texttt{RoBERTa-Sent-LU}                   & 0.65           \\

\textbf{Emotion Detection}                   &                  \\
\texttt{DistilBERT-EmoWoZ-FC}~\cite{huang2022distilbert}   & 0.56   \\
\texttt{DistilBERT-EmoWoZ-LU}          & 0.60           \\

\texttt{DistilRoBERTa-Emo-FC}~\cite{hartmann2022emotionenglish}    & 0.40 \\
 \texttt{DistilRoBERTa-Emo-LU}     & 0.44           \\

\textbf{Dialog Breakdown Detection}     &                  \\
\texttt{DBD+LogReg}~\cite{bodigutla-etal-2020-joint}  & 0.71    \\
\textbf{Rule-Based Approach}    &                  \\
\texttt{Keyword Matching}    & 0.41           \\
\textbf{LLM-based ICL Approach}      &                  \\
\texttt{GPT-4o-zero-shot-LU}     & 0.67\\
\texttt{GPT-4o-two-shot-LU}  &0.75	\\
\texttt{Llama-3.1-405B-zero-shot-LU}   & 0.63\\
\texttt{Llama-3.1-405B-two-shot-LU}   & 	 0.75\\
\texttt{GPT-4o-zero-shot-FC}     & 0.86\\
\texttt{GPT-4o-two-shot-FC}  &0.84	\\
\texttt{Llama-3.1-405B-zero-shot-FC}   & 0.83\\
\texttt{Llama-3.1-405B-two-shot-FC}   & 	 \textbf{0.87}\\
\bottomrule
\end{tabular}
\caption{Comparison of UF detections performed on last user utterance (\emph{LU}) and full conversations (\emph{FC}).}
\label{tab:results}
\end{table}
\section{Qualitative Analysis}
\label{sec:analysis}
This section compares academic benchmarks with real-world data and qualitatively analyzes open-source and in-house methods.

\subsection{Academic Data vs. Real-World Data}
\label{sec:analysis:academic}

\begin{table}[]
\centering
\small
\resizebox{\columnwidth}{!}{\begin{tabular}{lrr}
\toprule
         & EmoWoZ & Internal data\\
\midrule
\# Dialogues & 11,438 & 555 \\
\# Unique tokens & 28,417 & 995 \\
Avg. tokens / user turn & 10.6 & 3.1 \\
Avg. user tokens/dialogue & 55.6 & 7.8 \\
\% Repeated Utt. (fuzzy) & 2.1\% & 8\% \\
\% Repeated Utt. (cosine) & 4\% & 9.6\% \\
\bottomrule
\end{tabular}
}
\caption{Internal dataset vs. EmoWOZ benchmark.}
\label{tab:dialog_features}
\end{table}

We compare our data to an academic benchmark for emotion detection in TOD systems, EmoWoZ~\cite{feng-etal-2022-emowoz}, built in a controlled lab environment through Wizard-of-Oz and crowdsourcing techniques.
 In contrast, our internal dataset derives from real user interactions with a deployed TOD system, resulting in differences in emotional complexity, dialogue flow, frustration triggers, and user behavior. A key distinction is that real-world data includes dialogues where unfulfilled tasks can have tangible negative effects, creating a more urgent and authentic environment than academic benchmarks.

Through our analysis, which included two experts analyzing hundreds of conversations of both datasets, we observe several patterns indicating differences between them:
\begin{enumerate*}[label=\roman*)]
    \item different frustration triggers;
    \item real-world urgency vs lenient lab environment;
    \item number of user requests for human assistance;
    \item user familiarity with the system.
\end{enumerate*}

Frustration in our data stems from task-specific issues, like unavailable appointment times and the system's persistence with non-preferred options, while in EmoWoZ, it arises from a broader range of factors like misunderstandings or delays. Real users aim to complete tasks in real-time, so system failures lead to immediate dissatisfaction. In contrast, users in a lab environment face less pressure to complete tasks and can often move on to another goal within the same dialogue, giving the system a chance to recover or simply accepting the mistake.

Moreover, requests for human assistance are frequent and explicit in real-world, while almost non-existent in the EmoWoZ data. While this is understandable in academic benchmarks, where dialogues are acquired through WoZ techniques, such requests are nonetheless an important part of deployed TOD systems and should be looked at through the lens of potential frustration.
Finally, we observe cases where user familiarity with the capabilities of the system impacts their ability to fulfill their tasks without being frustrated. Such scenarios are not possible in academic benchmarks, as users in the lab are given specific instructions on how to interact with the system \cite{kim2024revealinguserfamiliaritybias}.

As shown in Table~\ref{tab:dialog_features}, frustration in real-world data is frequently conveyed through context and repetition rather than explicit wording. This aligns with the findings of qualitative analysis presented above. 
User responses are typically concise and to the point due to the nature of human-system interactions, with an average token count per user utterance reflecting this directness. Also the choice of words tends to be simpler and with less variability as indicated by a lower number of unique tokens present in the dialogues.

\subsection{Open-Source vs. In-House Methods}
\label{sec:analysis:qualitative}


\begin{figure*}[ht!]
 \hspace*{-0.7cm}
   \includegraphics[width=17cm, height=5cm]{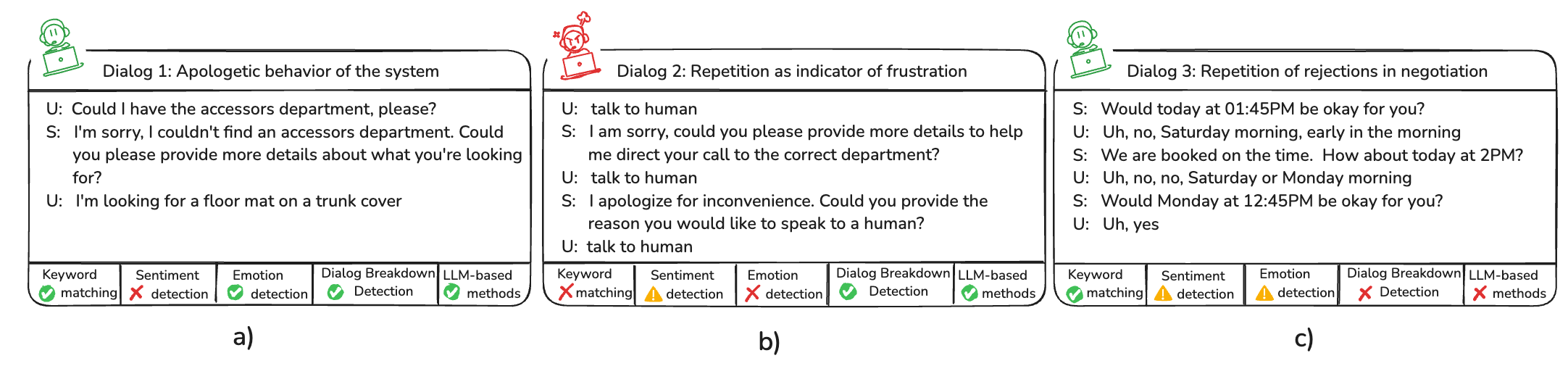}
  \caption{ 
    a) Negative sentiment prediction influenced by the apologetic behavior of the system
  b) Frustration caused by the system's failure to transfer the user to live agent. Yellow exclamation sign indicates that the example has been correctly classified by the FC-based sentiment approach only.
  c) Non-frustrated repetition of rejections in the process of time slot negotiations. Yellow exclamation sign indicates that the example has been correctly classified by LU-based and incorrectly by FC-based sentiment/emotion detection methods.  
}
\label{fig:dialog_examples}
\end{figure*}


We observe a limitation of sentiment and emotion detection methods in presence of system's apologetic behavior (e.g., responses containing phrases like \emph{``I'm sorry''} or  \emph{``I apologize''}), as shown on Fig.~\ref{fig:dialog_examples}a.
Such system utterances tend to dominate the overall perception of sentiment/emotion in the conversation, thereby diminishing the impact of expressed frustration in one or more user utterances.
Both the LLM- and DBD-based method are more robust to this phenomena.

Sentiment analysis and emotion detection approaches fail to recognize the repetition of users' requests, in cases where these requests were not initially met by the system,  as an indication of frustration~(Fig.~\ref{fig:dialog_examples}b).
On the other hand, LLM- and DBD-based methods more successfully capture such repetition pattern.

However, all of the approaches encounter difficulties in long negotiation scenarios (Fig~\ref{fig:dialog_examples}c), with a significant number of false positives arising from misinterpreting the negative sentiment and repetitions of users' rejections (e.g., responding ``no'' to system's question about user's availability at a certain time slot) as frustration. Moreover, emotion-based methods exhibit high sensitivity to interjections in the text, such as \emph{``uh''} or \emph{``ah''}.


Finally, we observe poor performance of LLM-based methods in short conversations, which often lack significant contextual information. These conversations typically consist of up to three turns.

\section{Discussion}
\label{sec:discussion}

\paragraph{User frustration detection is an important component in real-world TOD systems.}
We argue that, especially with the rise of popularity of conversational interfaces~\cite{mctear2017rise}, the task tackled in this study is essential for maintaining user satisfaction and engagement. 
However, as pointed out in Sect.~\ref{sec:analysis:academic}, current academic benchmarks for similar tasks are too sterile, as the dialogues were created in a controlled lab setting.
The differences between the real-world and academic benchmarks stem mainly from the real sense of urgency of fulfilling the task in the real world, while the simulated lab environment lacks the unpredictability and pressure of real-world scenarios.
Therefore, we call for additional attention to user frustration detection both from academia and industry practitioners.

\textbf{Frustration manifests in many ways, not limited to negative language.} As detailed in Sect.~\ref{sec:results}, our currently deployed keyword-based approach, that relies on identifying profane and negative language in user utterances, suffers from extremely low recall.
This, together with qualitative analysis presented in Sect.~\ref{sec:analysis:qualitative}, indicates that poor conversation handling does not always manifest as overtly negative language. Thus, while keyword-based methods may be inexpensive, they are inadequate for capturing the full range of frustrated scenarios.
On the other hand, our method fitted on the dialog breakdown detection features performs fairly well.
Although this approach is not directly comparable to the out-of-the-box or zero-shot methods, it shows how dialog breakdown relates to our user frustration task; features such as repetition and negation serve as strong indicators of frustration.

\textbf{General-domain emotion and sentiment models are insufficient for real-world TOD systems.} 
The poor performance might stem from the facts that the manifestation of frustration varies from person to person~\cite{bandura1973agression} and includes a wide variety of emotions, e.g., depression~\cite{berkowitz1989frustration}, thus going beyond the fixed set of emotions typically covered by pre-trained methods.
We hypothesize that their performance would increase if fine-tuned on domain data of a TOD system. However, such approach introduces maintenance overhead for systems operating across diverse and evolving domains. 
Further, a drop in performance when full conversations were used, indicates that such methods are over-sensitive to non-emotion-related text, as the overall emotion expressed by the user gets diluted by system's utterances.


\textbf{ICL with LLMs is an emerging method for frustration detection.} 
LLM-based methods outperform all baselines, suggesting that they capture both semantic- and dialog structure-related signals.
Moreover, similar performance of GPT-4o and Llama-3.1 demonstrates that our ICL prompt generalizes well across different LLMs. Another advantage of this approach is that it can be adapted to any domain as long as the domain is adequately described in the ICL prompt \cite{feng-etal-2024-affect}.


\section{Conclusions}
\label{sec:conclusions}

In this study, we investigated the feasibility of user frustration detection with out-of-the-box methods, including open-source sentiment and emotion detection, as well as deployed rule- and LLM-based methods.
We conclude that open-source methods are not fit for production TOD systems, likely due to the nature of the data, which is vastly different from real-world data, they were trained on.
Moreover, we find an LLM-based approach promising, as it tends to capture both emotion and potential dialog breakdowns, thus significantly outperforming other methods.
Future work encapsulates a promising direction of multi-modal (speech + text) methods for user frustration detection~\cite{ang2002}.
Finally, we aim to expand our detection across multiple user calls, therefore creating a user profile that can help with frustration detection.

\bibliography{custom}

\appendix

\section{In-Context Learning Prompt}\label{app:prompt}

Fig. \ref{fig:iclprompt} shows the different components of our in-context learning prompt for user frustration detection with Large Language Models (LLMs). The task definition and domain description are human-generated, and the conversation history is a variable which is sample-dependent. We also attach output instructions which inform the LLM to respond in the desired format.

\begin{figure*}[ht!]
\centering
\small
\begin{tcolorbox}[width=\textwidth, colframe=black, colback=blue!10, title=Task Description (T)]
\textbf{In this task, you are given a conversation between a user and a task-oriented dialog system. Your goal is to determine if the user is frustrated during the conversation.}

User frustration is often expressed through negative emotions, such as anger, irritation, or dissatisfaction. Some cues indicating frustration include:
\begin{itemize}
    \item Profanity or abusive language directed at the system.
    \item Hostility or irritation toward the assistant.
    \item Direct expressions of frustration, such as complaints about the system's performance or the conversation itself.
\end{itemize}

\textbf{Frustration can also be more subtle and does not always involve negative language}. A user may become frustrated when the system is unable to handle the conversation effectively or help the user accomplish their task. This may lead to breakdowns in the dialogue, where the user either disengages or expresses a desire to stop the interaction.

To classify frustration, consider the following signs:
\begin{itemize}
    \item Repetition of requests or questions due to the system's failure to resolve the user's issue.
    \item Use of negation, where the user rejects the system's suggestions or responses.
    \item Long, unresolved conversations where the user’s task remains incomplete.
    \item The user's general dissatisfaction with the system’s responses, even without overt hostility.
\end{itemize}
\end{tcolorbox}
\vspace{1em} 
\begin{tcolorbox}[width=\textwidth, colframe=black, colback=green!10, title=Domain (D)]
\textbf{The conversation you are analyzing occurs in one of these two domains}:
\begin{itemize}
    \item Receptionist system responsible for transferring calls to the appropriate department or agent. The system is expected to ask clarifying questions to help find the correct target.
    \item Booking agent that is negotiating the time slot for an appointment. We expect some back and forth between the user and the system to find a slot that works well for the user.
\end{itemize}

\end{tcolorbox}
\vspace{1em} 
\begin{tcolorbox}[width=\textwidth, colframe=black, colback=orange!10, title=Conversation History (H)]
\textbf{CONVERSATION: \{$chat\_history$\} }
\end{tcolorbox}
\vspace{1em} 
\begin{tcolorbox}[width=\textwidth, colframe=black, colback=red!10, title=Output Instructions]
\textbf{Return a single number:}
\begin{itemize}
    \item \textbf{0} if the user is \textbf{not frustrated}
    \item \textbf{1} if the user \textbf{is frustrated}
\end{itemize}
\end{tcolorbox}
\caption{In-Context Learning Prompt for User Frustration Detection in Task-Oriented Dialog Systems. The context is comprised of the description of the task ($T$), the domain of the conversation ($D$) and the conversation history ($H$). Our prompt also includes output instructions to generate binary user frustration labels.}
\label{fig:iclprompt}
\end{figure*}


\section{Dialog Breakdown Detection Features}
\label{dialogbreakdownappendix}
Table~\ref{tab:hand_crafted_features} lists the set of hand-crafted features utilized in the baseline from the \emph{Dialog Breakdown Detection} domain. The original set is given in~\citet{bodigutla-etal-2020-joint}. In our experiments, embeddings leveraged in the calculation of cosine similarity are created by the MPNet-like model (\emph{all-mpnet-base-v2}) trained using the process described in~\citet{reimers-gurevych-2019-sentence}.

\begin{table*}[h!]
\centering
\adjustbox{max width=\textwidth}{%
\begin{tabular}{llllllr}
\toprule    
        Feature name & Computation methodology \\
\midrule
Semantic paraphrase of user's req. & MA of the cosine similarity between pairs of utterances $u_{t-1}$ and $u_{t}$  \\
 Semantic repetition of system's resp. & MA of the cosine similarity between pairs of utterances $s_{t-1}$ and $s_{t}$  \\
   Semantic coherence of user's req. and system's resp. & MA of the cosine similarity between utterances $s_{t-1}$ and $u_{t}$  \\
 Syntactic paraphrase of user's req. & MA of the Jaccard index between sets of words of utterances $u_{t-1}$ and $u_{t}$  \\
 Syntactic repetition of system's resp. & MA of the Jaccard index between sets of words of utterances $s_{t-1}$ and $s_{t}$  \\
  Syntactic coherence of user's req. and system's resp. & MA of the Jaccard index between sets of words of utterances $s_{t-1}$ and $u_{t}$  \\
  Length of user's requests & MA of the utterance length $u_{t}$ \\
  Length of system's response & MA of the utterance length $s_{t}$ \\
 Length of full conversation & Number of characters in observed dialog $H$ \\
 Number of turns in conversation & Number of pairs $(u_{t}, s_{t})$ \\
\bottomrule
\end{tabular}
}

\caption{The set of features used in \emph{Dialog Breakdown Detection} approach. \emph{MA} represents the moving average across the consecutive pairs in the observed dialog $H$.}
\label{tab:hand_crafted_features}
\end{table*}

\section{Ethical Considerations}
As our study relies on data gathered from real users, we take several steps towards ensuring users' rights, privacy, and fair use of their data, in accordance with the US law.
First, prior to their conversation with our TOD system, we obtain an informed consent on the recording of the conversation and using the recording for any types of advancements of our system.
Second, we ensure privacy by performing anonymization of any potentially identifying user information.
Finally, we do not report any real user data in this paper.
Reported examples are manually augmented or rephrased in a way that preserves the right context, while ensuring no user utterance exactly matches the original user utterance.

\end{document}